\documentclass[11pt,a4paper]{article}
\usepackage[hyperref]{emnlp2020}
\usepackage{times}
\usepackage{latexsym}

\usepackage{microtype}

\aclfinalcopy % Uncomment this line for the final submission
\def\aclpaperid{2244} %  Enter the acl Paper ID here

\usepackage{subcaption}
\usepackage{latexsym}
\usepackage{amsmath}
\usepackage{multirow}
\usepackage{tablefootnote}
\usepackage{comment}
\usepackage{amssymb}
\usepackage{url}
\usepackage{enumitem}
\usepackage{array}
\usepackage{diagbox}
\usepackage{makecell}
\usepackage{setspace}
\setlist{nosep}

\usepackage{color}

\usepackage{graphicx,calc}
\DeclareMathOperator*{\argmax}{arg\,max}
\usepackage{xspace}
\newcommand{\cnn}{\textit{CNN}\xspace}
\newcommand{\glove}{\textit{CNN-GloVe}\xspace}
\newcommand{\trans}{\textit{Transformer}\xspace}
\newcommand{\bert}{\textit{BERT}\xspace}
\newcommand{\bertnlu}{\textit{BERT}-NLU\xspace}
\newcommand{\bertse}{\textit{BERT}-NLU-SE\xspace}
\newcommand{\bertce}{\textit{BERT}-NLU-CE\xspace}

\title{Learning to Generalize for Sequential Decision Making}

\author{Xusen Yin, Ralph Weischedel, and Jonathan May \\
Information Sciences Institute \\
University of Southern California \\
  \texttt{\{xusenyin,weisched,jonmay\}@isi.edu} \\}
  
\date{}

\begin{document}
\maketitle

\begin{abstract}
We consider problems of making sequences of decisions to accomplish tasks,  interacting via the medium of language. These problems are often tackled with reinforcement learning approaches. We find that these models do not generalize well when applied to novel task domains. However, the large amount of computation necessary to adequately train and explore the search space of sequential decision making, under a reinforcement learning paradigm, precludes the inclusion of large contextualized language models, which might otherwise enable the desired generalization ability. We introduce a teacher-student imitation learning methodology and a means of converting a reinforcement learning model into a natural language understanding model. Together, these methodologies enable the introduction of contextualized language models into the sequential decision making problem space. We show that models can learn faster and generalize more, leveraging both the imitation learning and the reformulation. Our models exceed teacher performance on various held-out decision problems, by up to 7\% on in-domain problems and 24\% on out-of-domain problems. 
\end{abstract}

\section{Introduction}

We make many decisions as we interact with the world. When we are rewarded (respectively, punished), we learn to modify not only the proximal cause of the stimulus but the chain of decisions leading up to it, to encourage (respectively, discourage) future similar results. This process naturally is the paradigm of \textit{Reinforcement Learning} (RL). Policy-based learning seeks to find good estimates for $Q(s, a)$, a function that returns the expected cumulative reward (known as a \textit{Q-value}) if action $a$ is chosen at state $s$. A desirable property of methodologies to learn $Q$ is their ability to \textit{generalize} such that an appropriate action can be taken when encountering a previously unseen state. 

Recent advances have shown strong evidence of generalization in spatiotemporal modalities such as robotic manipulation \cite{xu2018neural}, video games \cite{Tessler2017ADH}, and autonomous navigation \cite{Zhu_2017}. However, in the modality of language, there is less work applying generalization approaches to decision making. 
Useful applications of sequential decision making language models are
personal assistants that proactively anticipate client needs;
anti-phishing mediation agents that waste a would-be thief's time with relevant but non-helpful responses;
and investigative journalist assistants that determine what to read, whom to contact, and what questions to ask to create a revelatory news report.

Neural reinforcement learning (RL) training approaches, such as those used to play action video games \cite{mnih-atari-2013}, have potential applicability in language-based decision making due to their ability to learn to navigate adversarial or exploratory scenarios. Naturally, the generalization and background knowledge capability afforded by large contextualized language models such as \bert \cite{devlin-etal-2019-bert} may be applicable as well. A useful virtual world proxy in which to explore these approaches' applicability is that of \textit{text adventure game playing}. In a text adventure game, a player is immersed in an environment by reading textual descriptions of a scene and issuing natural language commands to navigate inside the scene. The player discovers and interacts with entities and accomplishes goals, while receiving explicit rewards for doing so. 

Learning to play text games is a useful pursuit because it is a \textit{convenient} proxy for the real world cases cited above. Unlike these, plentiful data for numerous games exist, an endless supply of games can be constructed, and text games have built-in reward functions,  making them suitable for RL. This class of problems is also useful because it is \textit{challenging}: after exposure to a \textit{family} of games that explore the same topic and have similar gameplay (e.g., games involving cooking a specified recipe), human players perform nearly perfectly on additional games, but computer models struggle. 

Why is this? Humans quickly understand the situation they are placed in and can make rational decisions based on trial-and-error and life experience, which we can call \textit{commonsense knowledge}. Knowing \textit{a priori} that, e.g.,  a \texttt{closed} door should be \texttt{open} or that it is helpful to \texttt{light a lamp} in a \texttt{dark dungeon} allows (human) players to learn  faster. Even though these games have the complexity of finite-state machines, computer models cannot learn to play them well. The problem appears to be due to a lack of generalization caused by a lack of commonsense. To a computer model, considering whether to \texttt{fry} using a \texttt{fridge} is no more ludicrous than considering whether to \texttt{fry} using a \texttt{plate} (which, to an untrained human cook, may be plausible, though is certainly not a good idea). Both actions can be discouraged by negative reinforcement, but a human only needs to learn not to do the latter.   Furthermore, a computer player learning that one can \texttt{chop carrots} with a \texttt{knife} may not generalize that one can \texttt{chop celery} the same way, but a human surely will.

There is existing work in learning to play text games with RL 
\cite{D15-1001,DBLP:journals/corr/abs-1806-11525,D14-1181,DBLP:conf/nips/ZahavyHMMM18,Yin2019ComprehensibleCT,DBLP:journals/corr/abs-1905-09700} but the standard pattern of incorporating large language models such as \bert \cite{devlin-etal-2019-bert} has not yet been seen in current literature. It turns out that this integration is not trivial. Most models that use \bert and its ilk predominantly apply their results to \textit{supervised learning} tasks that have training data with ground truth \cite{zellers-etal-2018-swag,wang-etal-2018-glue} or at least, in the case of generation-based tasks like dialogue and translation, a corpus of desirable output to mimic \cite{wolf2019transfertransfo,imamura-sumita-2019-recycling}. For tasks suited to RL such as the exploration of and interaction with a world, there is no true target or even, initially, a corpus, and thus learning can only proceed iteratively via, e.g., exploration-exploitation \cite{mnih-atari-2013}, which requires millions of training iterations to converge \cite{Yin2019ComprehensibleCT,DBLP:journals/corr/abs-1708-00133,mnih-atari-2013}. Integrating this process with the additional overhead of fine-tuning a large model like \bert leads to an impractical slowdown: for the experiments considered in this work, the baseline models that use \cnn require a little more than three weeks to train on an Nvidia P100 GPU-equipped machine. Using the same models on the same tasks run for the same number of iterations on the same hardware while fine-tuning a 12-layer \bert model would take more than two \textit{years}.

\begin{figure}[t]
\centering
    \includegraphics[width=0.45\textwidth]{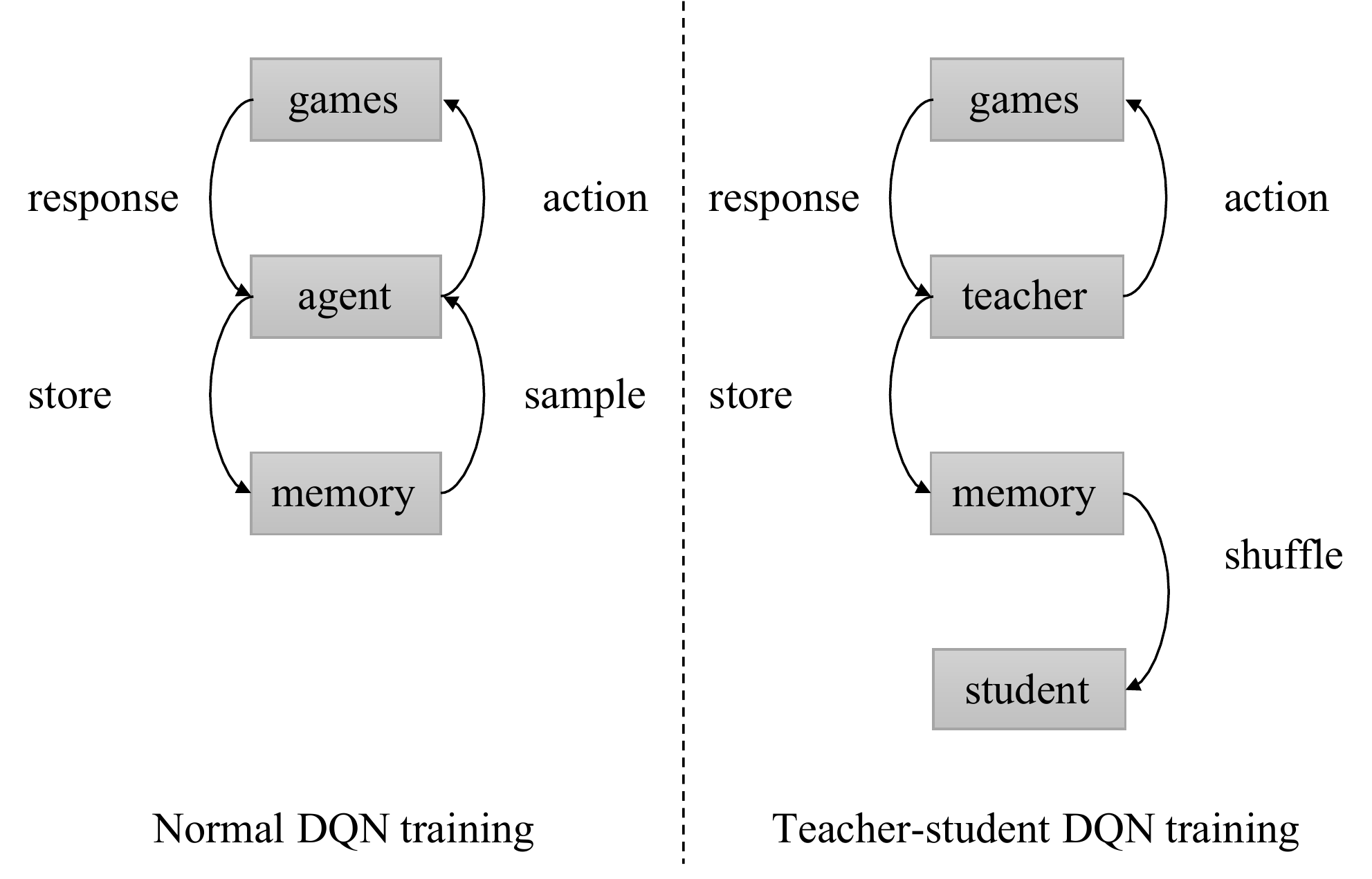}
    \caption{Comparison of the training process between DQN (left) and teacher-student DQN (right). In regular DQN training, the agent plays games to collect partial game-playing into memory, for later sampling and training. In the teacher-student training method, the teacher---a well-trained agent---generates a partial play curriculum. Student agents use that curriculum only to train their models.}
   \label{fig:teacher-student}
\end{figure}

In this work, we compare different previously used representation models for deep RL through an imitation learning method that first trains a light-weight teacher using exploration-exploitation, and then uses that trained model to train a more heavy-weight student model. This dramatically decreases the amount of training time needed to learn.  Moreover, we devise a means of casting an RL problem into a supervised learning paradigm, allowing better exploitation of large contextualized language models. In so doing, we show that agents can benefit from both the imitation learning and the reformulation, converging faster than other models, and exceeding teacher performance by 7\% and 24\% on both in- and out-of-domain problems, despite the limited search space.

The novel contributions of this work are:

\begin{itemize}
    \item We develop a teacher-student model training method for sequential text-based decision making problems, enabling the efficient incorporation of heavy-weight external information models.
    \item We develop a method for casting student RL model training in the same form as a supervised Natural Language Understanding task, enabling solutions to those tasks to be applied to sequential decision making.
    \item We evaluate our methods on in-domain and out-of-domain text game data sets, extrinsically and intrinsically demonstrating the effect of external commonsense knowledge and generalization at improving model abilities.
    \item We release our data, models, and code for documentation, replicability, and to enable subsequent improvements.\footnote{\url{https://github.com/yinxusen/learning_to_generalize}} 
\end{itemize}

\section{Background: Reinforcement Learning for Text Games}
\label{sec:rl}

The core approach of Deep Q-Networks (DQN)  as described by \newcite{google-atari} is to build a \textit{replay memory} of partial games with associated scores, and use this to learn a function $f_{DQN}: (S, A) \to \mathcal{R}$, where $f_{DQN}(s,a)$ is the Q-value obtained by choosing action $a \in A$ when in state $s \in S$; from $s$, choosing $\argmax_{\tilde{a} \in A} f_{DQN}(s,\tilde{a})$ affords the optimal action policy and this is generally used at inference time.\footnote{There are various methods to choose actions according to policies, but for exposition, the greedy method is the most representative and straightforward one. See Section~\ref{sec:eval} for details.}

The DQN, which predicts $Q$ given a (state, action) pair, is trained in an exploration-exploitation method known as \textit{$\epsilon$-greedy} \cite{google-atari}: first, the agent plays the game stochastically, generally guided by $f_{DQN}$, but with probability $\epsilon$ choosing a random action instead. 
The hyperparameter $\epsilon$ usually decays from 1 to 0 during the training process. 
As the agent plays, it collects partial play samples $(s, a, r, s')$, denoting taking action $a$ at the game state $s$, and the immediate reward $r$ plus the next state $s'$ reached for doing so, into a replay memory. The DQN is then improved by sampling from the replay memory, and reducing loss between $f_{DQN}$ and the true $Q$, which is estimated as 
\begin{equation} \label{eq:rlq}
Q(s, a) = r + \lambda \max_{a'} f_{DQN}(s', a').
\end{equation}
Square error loss is minimized to improve $f_{DQN}$ along the gradient: 
\begin{equation} \label{eq:sqer}
l_{dqn} = \|f_{DQN}(s, a) - Q(s, a)\|^2.
\end{equation}
The improved DQN is used to collect more replay data as the process iterates, as depicted in Figure \ref{fig:teacher-student} (left). 

Equation~\ref{eq:rlq} shows that at every step, for every sampled state, we can only estimate loss for a single state-action pair; we do not have the $r$ or $s'$ for actions other than $a$. The models eventually converge, but only after millions of training steps \cite{mnih-atari-2013,Yin2019ComprehensibleCT}. 

\section{Teacher-Student DQN Training}
\label{sec:teacher-student-training}

After running DQN training as described in Section~\ref{sec:rl} for some time, our well-trained agent, which we call the \textit{teacher}, can provide Q-values for a set of actions $A$ at every step. We can then collect lots of (state $s$, action set $A$, Q-table $Q$) game-play tuple data into a \textit{curriculum pool} by repeatedly playing the game and obtaining $f_{DQN-\mbox{teacher}}(s, a)$ for all $a \in A$. 
We now use that data to train a new agent $f_{DQN-\mbox{student}}$ (the \textit{student}), using the same DQN approach described in Section~\ref{sec:rl}.  However, unlike in the previous DQN scenario, the curriculum pool now contains Q-values for all of $A$ at each state.\footnote{It would not be viable or helpful to collect $Q$-values exhaustively during the $\epsilon$-greedy phase because of the poor initial estimates of $f_{DQN}$.}  We can train \textit{all actions at one step} for each trajectory since we have Q-values for all actions. Thus the loss is 
\begin{equation}
   l_{se} = \sum_{a\in A}\|f_{DQN}(s, a) - Q(s, a)\|^2 / \|A\|, 
\end{equation}
and the learning procedure is as depicted on the right side of Figure~\ref{fig:teacher-student}.

The main disadvantage of teacher-student learning is that in the student phase, the search space is bounded by that of the curriculum pool generated by the teacher agent. While a student model can generalize based on the curriculum pool's data, it cannot explore any more of the search space. On the other hand, student learning is much faster than teacher learning. The experience replay pool does not need to be repeatedly generated, and many more loss gradients can be calculated all at once. We will explore several architectures and configurations that take advantage of this speed.

\subsection{State Representations}

A fundamental parameter that must be specified is the input signal used to form the game state $s$ and how it is encoded. For action video games, this generally consists of a sequence of images from the game display. We use a history of system description-player action sequences for text games, which we call a \textit{trajectory}.  We consider the following representation architectures for the trajectory, some of which are only possible to use in the significantly faster student learning scenario:

\textbf{\cnn}. 
While much work applied to text games uses LSTMs \cite{lstm-original} to represent the trajectory \cite{D15-1001,ammanabrolu-riedl-2019-playing,DBLP:journals/corr/abs-1806-11525,DBLP:conf/cig/KostkaKKR17,DBLP:journals/corr/abs-1805-07274}, we favor \cnn encoders with position embeddings, which are faster to train than LSTMs \cite{DBLP:conf/nips/ZahavyHMMM18,Yin2019ComprehensibleCT,D14-1181}. This encoder is the only representation that is fast enough for training the teacher model, given the fact that the trajectory length is usually much longer than a single sentence or paragraph. We also experiment with it as a student model trajectory representation. This baseline \cnn encoder uses randomly initialized word embeddings that are fine-tuned during training. This encoder has one layer, with 32 convolutional filters for each of size 3--5 \cite{D14-1181}.

\textbf{\glove}. The \glove encoder is identical to the \cnn encoder except for the use of GloVe \cite{pennington2014glove} for word embeddings; these are not fine-tuned.

\textbf{\trans}. We use the \trans \cite{NIPS2017_7181} architecture configured in the same way as the \bert-base uncased model with 12 layers \cite{devlin-etal-2019-bert}, but with all weights randomly initialized. This model will serve as a comparison with the following model.

\textbf{\bert}. We use the \bert-base uncased model with 12 layers. This model has the same architecture as \trans but is initialized with \bert weights \cite{devlin-etal-2019-bert}. 

We use a max-pooling layer over the output of \cnn as the encoded state in the same way that we do with \glove, while for \trans and \bert, we use the pooling output from the \texttt{CLS} token as the encoded state. All encoded states from different encoders are passed through a dense linear layer of 32 dimensions to ensure the encoded state is of equal size across models.

We use \bert's provided Byte-Pair Encoding \cite{Sennrich_2016} sub-word tokenizer and vocabulary with 30,522 tokens for \cnn, \trans, and \bert. For \glove, we use the GloVe 6B model with 400,000 tokens and the TreeBank word tokenizer from NLTK \cite{10.3115/1118108.1118117} since GloVe embeddings are pre-determined and not compatible with BPE. We use a zero vector as the padding token and average of all word embeddings as the unknown token for \glove. \cnn uses a word embedding size of 64, while for \glove and \bert, we use the pre-trained word embedding size, i.e., 50 dimensions for \glove (we choose this dimension because it is close to our \cnn) and 768 for \bert (so does \trans).

\subsection{Action Representations}
\label{sec:drrn}

\begin{figure}[t]
\centering
    \includegraphics[width=0.45\textwidth]{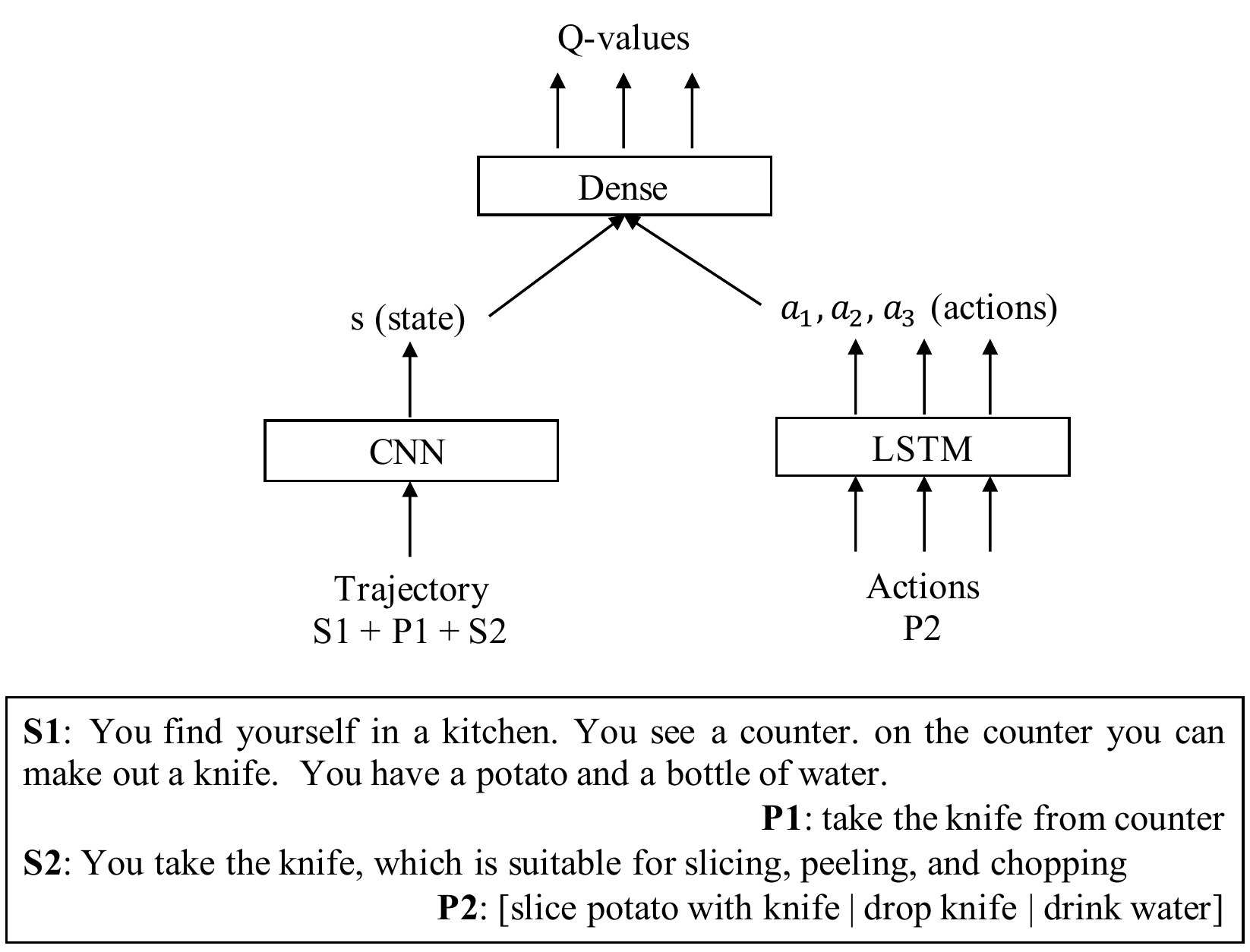}
    \caption{The architecture of the DRRN model. Trajectories and actions are encoded by a \cnn (in this case) and an LSTM into state and action representations, respectively, followed by a dense layer to compute the Q-values. On the bottom, we show a truncated example of dialogue from a text game in the cooking genre, with S1 and S2 representing the system's descriptions, and P1 showing the player's first actions in response to S1. S1 + P1 + S2 is an example of a trajectory. P2 shows a set of admissible actions.}
   \label{fig:archi-drrn}
\end{figure}
 
A consequence of learning to play different games is that actions differ from one game to another. Vanilla DQNs, introduced by \cite{google-atari}, are incompatible with this modification since they presume a predefined finite and consistent action space, such as the directions and push buttons of a joystick. 
Additionally, vanilla DQNs presume no semantic relatedness among action spaces. In text games, by contrast, it would make sense for, e.g., \texttt{open the door} to be semantically closer to \texttt{shut the door} than to \texttt{dice the carrot}. In our experiments, we assume the action set for a test game may be unknown at training time, and that actions may have some interrelatedness.\footnote{This is itself still a simplification, as many text games allow \textit{unbounded} action generation. We leave this problem for future work.} 
We thus represent actions using  Deep Reinforcement Relevance Networks (DRRN)  \cite{P16-1153}, a modification of the standard DQN, as shown in Figure \ref{fig:archi-drrn}. Actions are encoded via an LSTM \cite{lstm-original} and scored against state representations by inner products and an extra dense layer. In preliminary experiments, we found that LSTMs worked better than \cnn on the small and similar actions in our space, such as \textit{take yellow potato from fridge} and \textit{dice purple potato}. We limit these actions to a maximum length of 10 tokens. We use DRRNs in both teacher and student scenarios.

\subsection{Game-Playing as Sequential Natural Language Understanding Tasks}
\label{sec:nlu}

\begin{figure}[t]
\centering
    \includegraphics[width=0.4\textwidth]{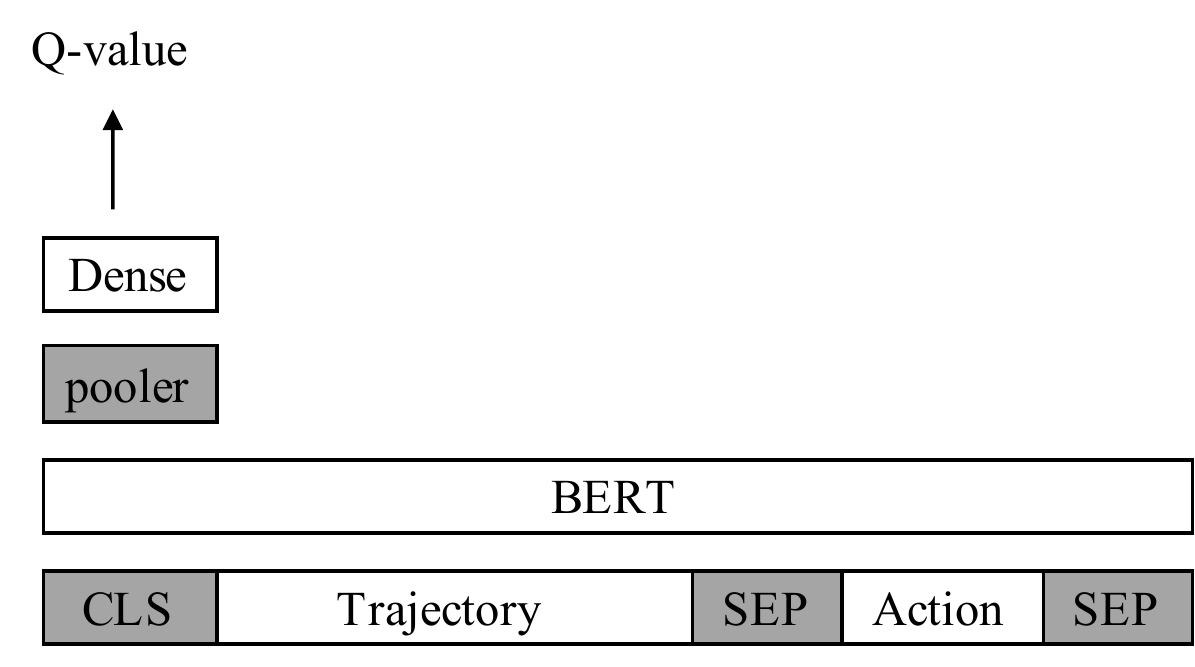}
    \caption{The architecture of the \bertnlu models (\bertse shown). Trajectories and actions are concatenated together as the input to \bert, followed by a dense layer over the pooling output of the CLS token to compute the Q-value. Unlike the DRRN model, we need to concatenate a trajectory with an action to compute a Q-value, which increases computation time.}
   \label{fig:archi-nlu}
\end{figure}

Large contextualized language models built on the \trans architecture such as \bert and GPT \cite{radford2018improving}  have been used in supervised multiple-choice Natural Language Understanding (NLU) tasks.
While we have so far encoded trajectories and actions separately in the DRRN formulation of DQN (Section~\ref{sec:drrn}),
NLU task architectures commonly encode context and hypothesis together, using a dense final layer to obtain scalar confidence in the hypothesis being the correct result of the context. This is then trained (with a cross-entropy loss) across all hypotheses for that context. 
By considering trajectories as context, actions as hypotheses, and $\argmax_{\tilde{a} \in A} f_{DQN-\mbox{teacher}}(t, \tilde{a})$\footnote{Henceforth we abuse notation and replace $s$, the state, with $t$, the trajectory used to represent it} as a label for trajectory $t$ from the curriculum pool, we may easily switch to  this framework, now minimizing a standard cross-entropy loss in place of DQN student learning. We call this model \bertce. At evaluation time, the model chooses an action to take given a trajectory, but we are no longer explicitly learning a new $Q$-function other than simply learning a preferred choice.

Of course, having an existing $Q$-table from the teacher model, we may instead replace the cross-entropy loss with the familiar mean squared error loss (Section~\ref{sec:teacher-student-training}). This model, which we call \bertse, operates the same way as \bertce, but the values associated with each (trajectory, action) pair are once again regarded as $Q$-values. Figure~\ref{fig:archi-nlu} depicts the architecture of \bertse; \bertce is identical except the output is not explicitly intended to be a $Q$-value.

While most NLU tasks like SWAG \cite{zellers-etal-2018-swag} or ARC \cite{clark2018think} have no more than five hypotheses to choose from, even artificially constrained text-based games may have hundreds of potential choices. To make training feasible for text games, given each trajectory, we randomly sample three possible actions, along with the teacher model's most favored one.
At evaluation time, the model can choose from all admissible actions.

\section{Games and Evaluation Methodology}

Unlike most video- or text-game-playing work \cite{mnih-atari-2013,DBLP:conf/nips/ZahavyHMMM18,Yin2019ComprehensibleCT,D15-1001} which incrementally learns to play games through RL approaches and reports results on those same games, we evaluate on games that are not seen during learning. Our games are generated from the TextWorld platform \cite{textworld-a-learning-environment-for-text-based-games}, which procedurally generates a wide variety of game variants with different maps, objectives, actions, threats, and background text, given user-supplied inputs. The platform provides the set of \textit{admissible} actions, i.e., legal actions available at each state of each game. There are between 10--100 of these actions depending on the context. 

\subsection{Training and Evaluation Data}
\label{sec:traindata}

We use the games released by Microsoft for the \textit{First TextWorld Problems}\footnote{\url{https://www.microsoft.com/en-us/research/project/textworld}} competition for our training set and an evaluation set of unseen but in-domain games. The competition provides 4,440 cooking games generated by the TextWorld framework. The goal of each game is to prepare a recipe. The action space is simple, yet expressive, and has a moderately large, though domain-limited, vocabulary. 
One example presents a recipe with directions such as \texttt{fry the pork chop,  chop the yellow bell pepper,   fry the yellow bell pepper,   prepare meal}. To succeed, the player must find and take the items in various rooms and containers, use the correct cooking implements, and not spoil the ingredients, e.g., by \texttt{fry}ing an \texttt{apple} that has already been fried. We hold out 444 of these games as an in-domain test set.

To evaluate our models' ability to generalize beyond their training domain, we also evaluate on an out-of-dojain set, comprising 208 newly generated games\footnote{\url{https://github.com/microsoft/TextWorld\#usage}} in a treasure-hunting genre. These have quite different actions, objects, and goals from cooking games. They generally require the player to navigate around rooms, find a specific object, and take a specified action with the entity, e.g., picking up a \texttt{key} and \texttt{insert}ing into a \texttt{gate}'s lock in a different room to \texttt{unlock} it. These games have little vocabulary overlap with any cooking games apart from basic commands like \texttt{take} and \texttt{drop}.

\subsection{Evaluation}
\label{sec:eval}

We report scores on each test set as a percentage of the possible total score. Each game has 1--6 points available. At evaluation time, we play each game twice, stopping after the sooner of 100 steps, game completion, or game failure, and consider each play independently. Scores can vary because each gameplay uses an initial knowledge graph map construction built via random walks \cite{ammanabrolu-riedl-2019-playing} and because confidence bound is learned per action \cite{yin2019learn}, such that at evaluation time, lower-confidence actions are chosen with more stochasticity. An agent taking purely random walks (a low-bar baseline) scores 14\% on the in-domain test and 16\% on out-of-domain.

We train the teacher agent for 10 million steps on the 3,960 training games in the cooking domain, using deep Q-learning described in Section~\ref{sec:drrn}. We use a curriculum training schema \cite{yin2019learn} to train our teacher model.
During training, each 5,000-step checkpoint takes 25 minutes on a single P100 GPU. We decay $\epsilon$ from 1 to 0 during training. The teacher agent scores 70\% on in-domain test and 33\% on out-of-domain test.  

\section{Student Experiment Results}

Having trained a teacher model using DQN and allowing unlimited exploration of the game space, we now experiment with several student learning approaches. Relative to the teacher model, the students are constrained to explore using data generated from the trained teacher model. This restriction limits their ability to search but enables much faster training and, consequently, richer models. All student models are trained for 500,000 steps of 32 (trajectory, action) pairs per batch, saving checkpoints every 5,000 steps and generating results for in- and out-of-domain test sets. Running on a single P100 GPU, all \trans-based models take 75-80 minutes per 5000 steps, while \cnn-based models take 13 minutes. 

\subsection{Data Generation from Teacher Models}

We generate student curriculum pools from the trained teacher model by playing all Cooking-Train games in random order. Specifically, we play games with the teacher agent using $\epsilon$-greedy search (Section \ref{sec:rl}). We uniformly sample $\epsilon\in [0, 1]$ among different game playing episodes to increase the variety of trajectories exhibited to student learners. We collect the trajectory, all actions, and Q-values assigned to each action by the teacher model for each game playing step. In total, we collect 10 million instances of such tuples from the 3,960 Cooking-Train games.

\begin{figure}[t]
    \centering
    \includegraphics[trim=10mm 3mm 3mm 5mm,width=0.45\textwidth]{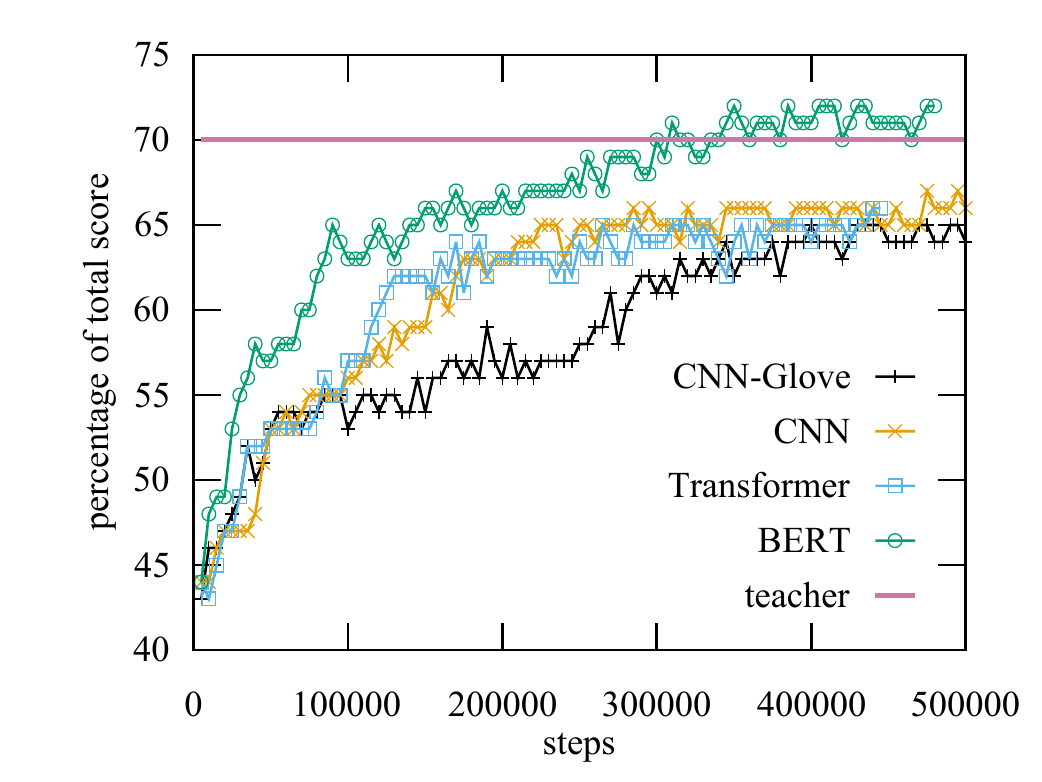}
    \caption{Comparing DRRN state encoders over in-domain test games. \bert converges faster and to a higher point (72\%) than other students (\cnn 67\%; \glove 65\%; \trans: 67\%) and the teacher (70\%).}
    \label{fig:chart-1}
\end{figure}

\subsection{In-Domain DRRN Results}

In Figure~\ref{fig:chart-1}, we compare student model learning to the teacher model's final position (horizontal line). We see that for many of the models, the trade-off to a potentially more sophisticated architecture is not worth the damage caused by limited exploration. As expected, our baseline model, \cnn, which is the same model used for teacher training, converges to 67\% of the total possible score at around 300,000 steps of training; the teacher model is at 70\%.
\glove, compared to \cnn, is even worse and converges more slowly. Even though \glove is equipped with pre-trained word embeddings, the student agent cannot benefit from it.

\trans performs comparably to \cnn, but \bert learns much more quickly than all other models, reaching 72\% of the total score on test games; 5\% higher than any other student models and somewhat better than the teacher model, which is an encouraging preliminary result.

\subsection{In-Domain NLU Results}
\label{sec:in-domain-nlu}

\begin{figure}[t]
    \centering
    \includegraphics[trim=10mm 3mm 3mm 5mm, width=0.45\textwidth]{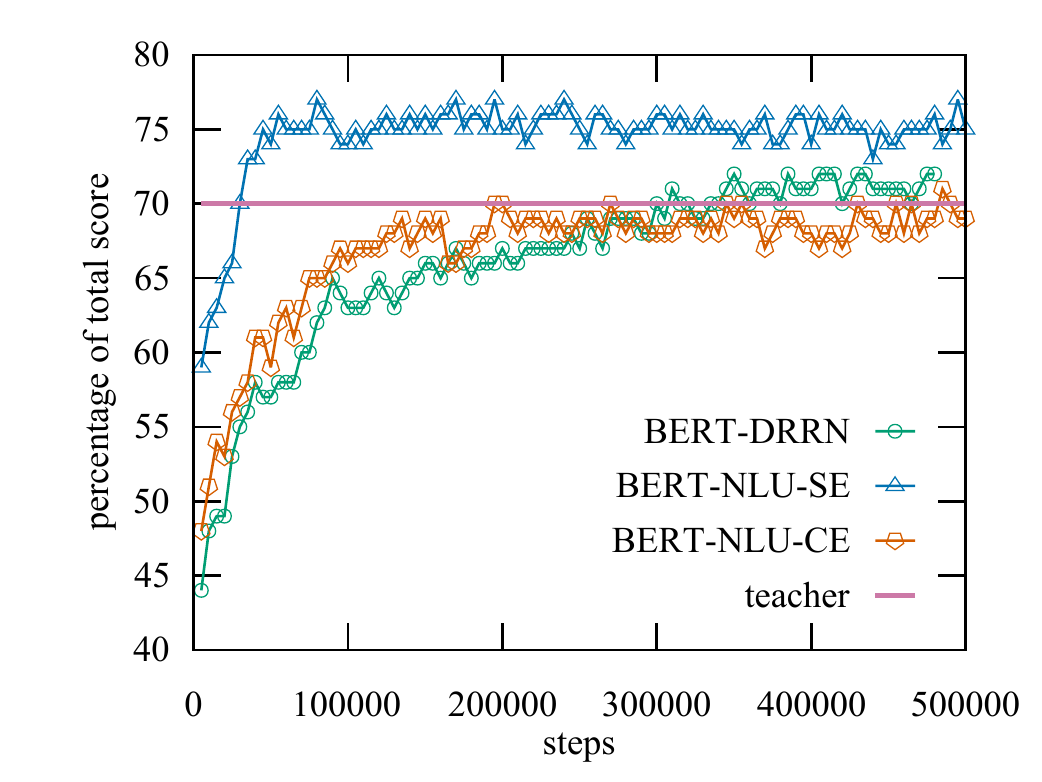}
    \caption{Comparing the use of \bert in a DRRN (Section \ref{sec:drrn}) or NLU (Section \ref{sec:nlu}) student model over in-domain test games. \bertse converges 7\% higher than the teacher model very rapidly, indicating quick generalization capability.}
    \label{fig:chart-2}
\end{figure}

\begin{table}[ht]
\centering
\begin{tabular}{lrrrrr}
\multirow{2}{*}{} & \multirow{2}{*}{GRD} & \multirow{2}{*}{UCB} & \multicolumn{3}{c}{Sampling} \\  
                        &                         &                         & t=0.01    & t=0.1   & t=1.0   \\
SE                      & 72                      & 77                      & 72        & 66      & 27      \\
CE                      & 67                      & 71                      & 65        & 65      & 66      \\

\end{tabular}
\caption{Comparing different inference methods for \bertnlu agents over cooking-Test (percentage of total possible score). We compare two \bertnlu agents, one trained with square error loss (SE) and one with cross entropy loss (CE). We use three different methods of choosing actions from policies: greedy (GRD), LinUCB (UCB) and sampling at three different temperatures.}
\label{tbl:choose-action}
\end{table}

\begin{table}[ht]
\centering
\begin{tabular}{lrr}
                         & SE   & CE   \\
\bert                 & 77 & 71 \\
\textit{no-init}              & 69 & 60 \\
\textit{freeze-all-but-pooler} & 26 & 25 \\
\textit{freeze-to-penultimate} & 61 & 54
\end{tabular}
\caption{Ablation study of the NLU method. We compare the best test-set evaluation scores (percentage of total) over the 100 epochs of training for each model trained with either square error loss (SE) or cross-entropy loss (CE).}
\label{tbl:ablation-nlu}
\end{table}

In Figure \ref{fig:chart-2} we explore the performance of the NLU-inspired architecture (Section~\ref{sec:nlu}).  The cross-entropy-based approach, \bertce, is the most similar to standard supervised NLU tasks and performs comparably to the DRRN teacher-student framework. However, \bertse, which directly regresses to the $Q$-function's value, quickly converges to around 77\% of optimal scores, 7 points higher than the teacher model.

Independent of the method for learning Q-values, we can choose between multiple methods to \textit{apply} the policy at inference time. We compare three frequently used methods---$\epsilon$-greedy, sampling, and LinUCB, a bandit feedback method \cite{Auer2003,Abe2003RL,NIPS2011_4417}---in Table~\ref{tbl:choose-action}. Following \newcite{yin2019learn}, we use $\epsilon=0$ for the $\epsilon$-greedy method. 
For the sampling method, we choose different temperatures over the Q-values. We follow \newcite{yin2019learn} for the LinUCB method. In Table~\ref{tbl:choose-action}, we ablate \bertce and \bertse training and five different inference approaches. The same Q-values are used for each setting using \bertce and for each setting using \bertse.  We find that models trained with square error loss and evaluated using sampling are highly sensistive to temperature; cross-entropy-trained models are fairly insensitive. However, both the $\epsilon$-greedy and the sampling methods perform worse than the LinUCB method.

We ablate the impact of fine-tuning \bert in Table~\ref{tbl:ablation-nlu}, showing what happens if we do not fine-tune except the pooler (\textit{freeze-all-but-pooler}), only fine-tune the last layer and the pooler (\textit{freeze-to-penultimate}), or fine-tune all layers (\bert). We also show the fine-tuned equivalent \trans that is not pre-trained (\textit{no-init}) for comparison. All settings fine-tune the 768-parameter last dense layer on top of \trans to compute Q-values. The {\it freeze-to-penultimate} allows the final Q-value layer, the {\it pooler}, and the last layer of \bert to train. In total, more than seven million parameters are trainable in {\it freeze-to-penultimate}. However, the performance still has a 16\% gap compared to the fully fine-tuned 110-million-parameter \bert models. This ablation study shows that the benefits coming from \bert can not be reproduced by simply using out-of-the-box \bert weights, which would speed up the training process, and underscores the importance of imitation learning.

\subsection{Out-of-Domain Results}

\begin{figure}[t]
    \centering
    \includegraphics[trim=9mm 4mm 9mm 5mm, width=0.48\textwidth]{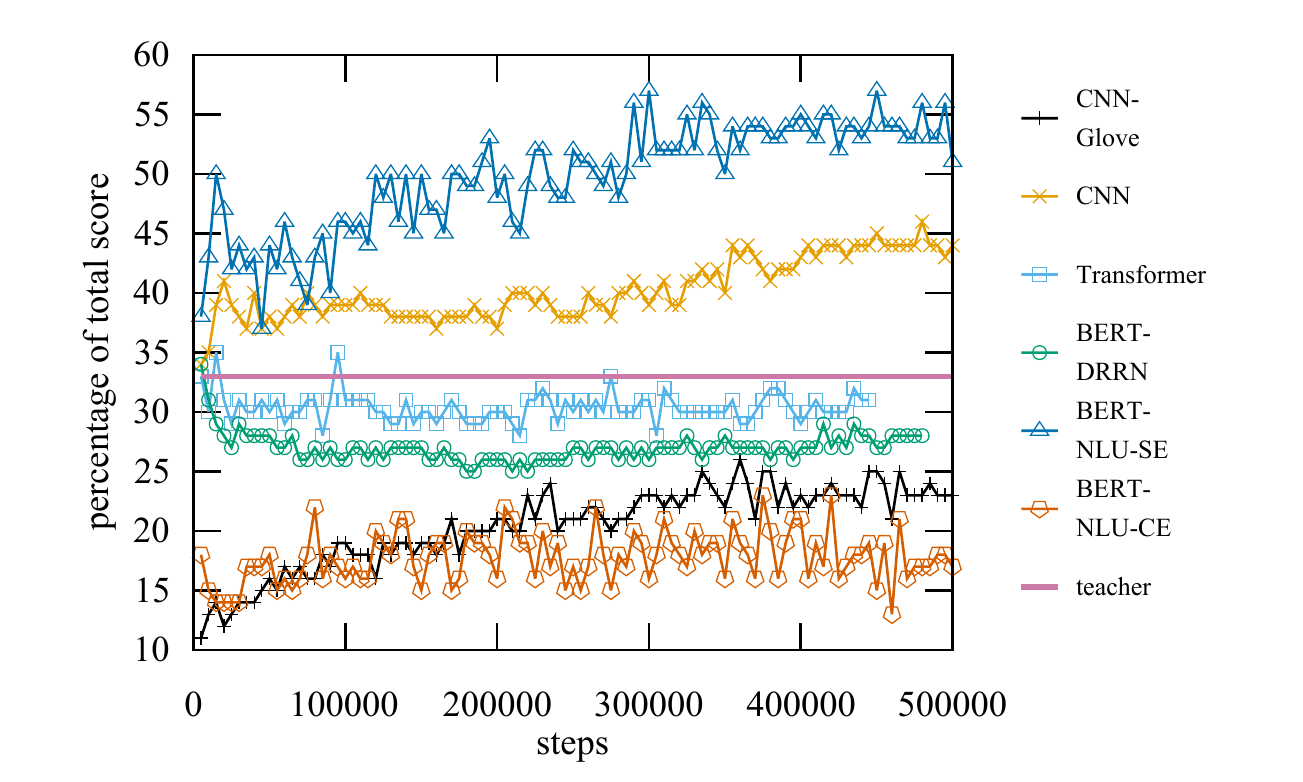}
    \caption{Comparing performance on out-of-domain test games. Only two agents exceed teacher results (by 24\% and 13\%, respectively). Analysis in Figure~\ref{fig:kl-d} shows that the superior performing \bertse benefits from external knowledge, while \cnn-DRRN mostly benefits from stochasticity.}
    \label{fig:chart-3}
\end{figure}

 Figure~\ref{fig:chart-3} shows the result of evaluating with out-of-domain games. These games have different goals and action sets from the training games, so it is possible during training to observe performance curve drops, an indication that the model is overfitting on the cooking game genre and not properly generalizing. Most of the DRRN student models exhibit some overfitting; only the \cnn model can learn somewhat well and exceeds the performance (46\%) of the teacher model (33\%). \bertse, the NLU-style architecture that fine-tunes \bert and is trained to directly estimate $Q$-values,greatly exceeds the teacher model's performance (57\%) on these games from an unseen genre.
 
 \section{Discussion}
 
 In this section we seek to understand the following:
 
 \begin{itemize}
     \item What extra information \bertse leverages compared to other DRRN models (Figures~\ref{fig:chart-1} and \ref{fig:chart-2});
     \item What generalization and extra information \bertse leverages on out-of-domain games, and why the \cnn student model performs better than expected on out-of-domain games (Figure~\ref{fig:chart-3}).
 \end{itemize}

A qualitative investigation of model performance on in-domain test sets shows that game failure arises when a model decides to prepare an ingredient improperly, (e.g., to use the \texttt{BBQ} instead of the \texttt{stove} to \texttt{fry}). Models initialized with \bert have fewer such failures, indicating that \bert provides background cooking knowledge, beyond what can be learned from the curriculum pool. Example gameplays and complete statistics on test games are provided in the Appendix.
 
A similar pattern is observed for out-of-domain tests. One test requires the player to use four different kinds of keys with matched locked containers. As the curriculum pool does not have any information relevant for this task, models without general background knowledge suffer. In the key/lock test game (a readout is in the Appendix), the teacher model repeatedly \texttt{unlock}s and \texttt{lock}s a single box, and \texttt{put}s and \texttt{take}s the same key without making progress. The \bertse model, however, can correctly open the sequence of containers. 
 
\begin{figure}[t]
    \centering
    \includegraphics[width=0.48\textwidth]{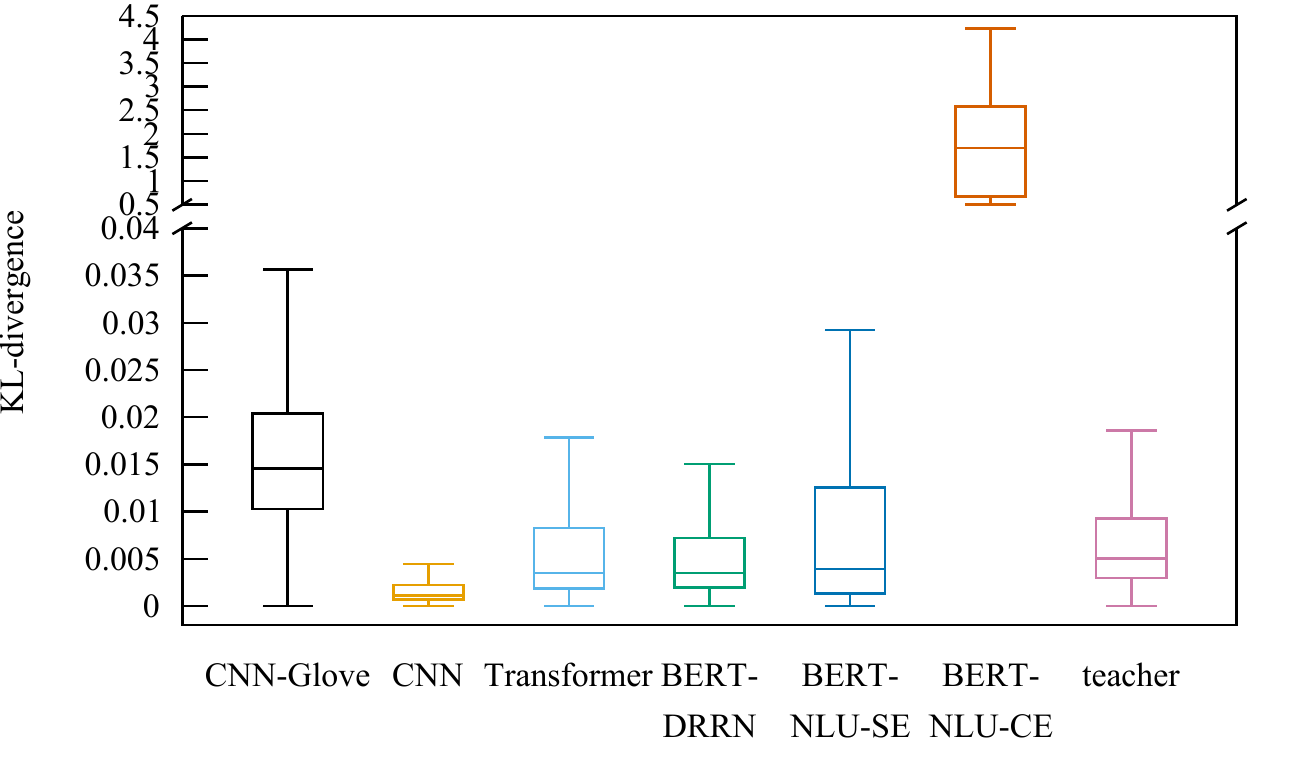}
    \caption{Boxplot of KL-divergence between Q-values and uniform distributions for agents over treasure hunting games, showing agent \textit{confidence}. A larger KL-divergence value indicates more confidence and thus a sharper distribution of Q-values. The \cnn student agent is the least confident agent, but this allows it to explore more; other models are more confident but are wrong. Only \bertse is both confident and correct, able to generalize well.}
    \label{fig:kl-d}
\end{figure}

Figure~\ref{fig:kl-d} provides more insight into model performance, including an explanation for the surprising success of the \cnn model. That figure shows the KL-divergence \cite{kullback1951} between a uniform distribution and the distribution formed from the $Q$-values (the categorical choice distribution for \bertce) at every step during every out-of-domain test, computed from the final point of each model. The \cnn model's distribution is closer to uniform than the others. As stochastic choices are made at test time when the action distribution is uncertain (see Section~\ref{sec:eval}), the \cnn model performs more exploration during the evaluation of Treasure hunting games. These games do not have failure cases like the in-domain test games, so there can be some benefit to stochasticity. The other models are more confident\footnote{particularly \bertce, which is trained to make very peaked decisions} and, except for \bertse, are generally wrong. This result indicates that equipped with the ability to generalize from \bert pre-training, \bertse has learned the skill of decision making, rather than the ability to memorize patterns. 
 
\section{Related Work}
Many recent works \cite{D15-1001,P16-1153,DBLP:journals/corr/abs-1805-07274,fulda2017affordance,textworld-a-learning-environment-for-text-based-games,DBLP:conf/cig/KostkaKKR17} on building agents to play text-based games apply DQNs \cite{google-atari} or their variants. Different aspects of DQN have been presented, such as action reduction with language correlation \cite{fulda2017affordance}, a bounding method \cite{DBLP:conf/nips/ZahavyHMMM18}, the introduction of a knowledge graph \cite{ammanabrolu-riedl-2019-playing}, text understanding with dependency parsing \cite{Yin2019ComprehensibleCT}, and the bandit feedback method for agent evaluation \cite{yin2019learn}.

However, previous work uses different games to evaluate, making it difficult to compare results comprehensively. With the TextWorld framework's availability, there is more and more work concentrating on the generalization ability of agents, which seldom appears in the video game playing domain. \newcite{DBLP:journals/corr/abs-1806-11525} work on generalization of agents on variants of a very simple coin-collecting game. The simplicity of their games enables them to use an LSTM-DQN method with a counting-based reward. \newcite{ammanabrolu-riedl-2019-playing} use a knowledge graph as a persistent memory to encode states, while we use a knowledge graph later on to make actions more informative.

The TextWorld competition has yielded a variety of works that use different approaches and methods:
\newcite{Yuan_2019} co-train a DQN with a question answering system for building new interactive machine reading comprehension tasks while creating agents to solve games.
\newcite{Madotto_2020} describe a non-RL method to learn agents, by first randomly playing on training games, then collecting all winning trajectories. By using these trajectories as training data, they manage to transform an RL problem into supervised learning. 
\newcite{Adolphs_2020} use an actor-critic framework and prune the action-space by using hierarchical RL and a specialized module trained on a recipe database to build better agents.
\newcite{Jain_2020} apply the action elimination method proposed by \newcite{DBLP:conf/nips/ZahavyHMMM18} on Zork to the cooking games.

For teacher-student training, \newcite{rusu2015policy} design a policy distillation method that trains different agents as teacher agents. Each of these teacher agents learns to play a single and separate game. Then they build one student learner that can be trained with a supervised learning method to distill the policy knowledge for multi-game playing. \newcite{DBLP:journals/corr/abs-1805-07274} also use teacher-student training for text-based games. However, our teacher-student training method is different: we use one teacher that can play multiple games to guide multiple student agents' learning processes.

\section{Conclusion}

We provide a recipe for integrating large contextualized language models and deep reinforcement learning, applying to sequential decision making and a demonstration on the proxy task of text games, showing dramatic improvements over the standard practice, particularly in out-of-domain held-out tests. We expect to apply this approach to various challenging real-world sequential decision scenarios, such as goal-directed dialogue and active information-gathering.

\section*{Acknowledgments}

This research is based upon work supported in part by the DARPA Machine Common Sense program, Contract N660011924032 through NIWC.  The views and conclusions contained herein are those of the authors and should not be interpreted as necessarily representing the official policies or endorsements, either expressed or implied, of DARPA or the U.S. Government. The U.S. Government is authorized to reproduce and distribute reprints for Governmental purposes notwithstanding any copyright annotation thereon.

\bibliography{anthology,emnlp2020}
\bibliographystyle{acl_natbib}

\cleardoublepage
\newpage

\appendix

\section{Appendix}
\label{sec:appendix}

This appendix contains comprehensive results of the models explored in this work on the two test sets. Table~\ref{tbl:teacher-and-random} shows the number of in-domain games won, lost due to incorrect handling of materials, and lost due to taking too many steps; the latter category is helpful, along with $Q$-table KL-divergence from a uniform distribution, in explaining the poor performance of overly deterministic \bertce, which fails very infrequently, but often gets stuck in a step loop instead. Table~\ref{tbl:teacher-and-random-th} shows the number of out-of-domain games won per model, as out-of-domain games cannot be lost other than by taking too many steps. We can subtract the numbers in this table from 416 to find how many games a model lost. The column `wins over teacher,' for all models other than the teacher, shows the number of games the model won that the teacher model did not. As all failures here were due to taking too many steps, stochasticity helps \cnn just as much as determinism hurts \bertce. Sections~\ref{sec:bert-th-readout} and \ref{sec:teach-th-readout} contain one complete out-of-domain game, as played by \bertse and the teacher agent. Sections~\ref{sec:bert-cook-readout} and \ref{sec:teach-cook-readout} show the same for a cooking game.

\begin{table*}[]
\centering
\begin{tabular}{lrrr}
             & Win ($\uparrow$) & Fail ($\downarrow$) & too many steps ($\downarrow$) \\
teacher      &    578         &       245         &   65     \\
\cnn         &    543         &       306         & 39       \\
\glove       &    525          &       317         &   46     \\
\trans       &    551         &       233         & 104      \\
\bert-DRRN   &    596         &       190         &  102     \\
\bertce &    572         &       87          &  229     \\
\bertse &    631         &       150         &  107     \\

\end{tabular}
\caption{Tally of in-domain test game outcomes for the models explored in this work, indicating which games are won, which games lead to a failure (by spoiling ingredients), and which end when the step limit (100) is reached. \bertce and \bertse fail on in-domain tests the least, indicating a better intuitive understanding of how to prepare recipes. Because \bertce is so confident (Figure~\ref{fig:kl-d}) it can get stuck in a deterministic loop; this is reflected in a larger number of  `too many steps' outcomes relative to \bertse.}
\label{tbl:teacher-and-random}
\end{table*}

\begin{table}[]
\centering
\begin{tabular}{lrr}
             & Wins  & Wins over teacher  \\
teacher      &    122         &       -        \\
\cnn         &    173         &       70        \\
\glove       &    85          &       24        \\
\trans       &    112         &       25        \\
\bert-DRRN   &    102         &       19        \\
\bertce &    56         &       5         \\
\bertse &    209         &       106        \\

\end{tabular}
\caption{Tally of out-domain test game outcomes for the models explored in this work, indicating which games are won and which games are won in excess of those won by the teacher. }
\label{tbl:teacher-and-random-th}
\end{table}

%\cleardoublepage
%\newpage
\section{\bertse on one Treasure Hunting Game}
\label{sec:bert-th-readout}
\setlength\parindent{0pt}
%\small
Who's got a virtual machine and is about to play through an life changing round of TextWorld? You do! Here is how to play! Your first objective is to pick up the keycard in the cubicle. Then, insert the keycard into the box within the cubicle's lock to unlock it. After you have unlocked the box, ensure that the box in the cubicle is open. Having pulled open the box, retrieve the type A latchkey from the box. After that, unlock the type A locker inside the cubicle. Then, ensure that the type A locker in the cubicle is open. After that, pick up the key from the type A locker. Then, unlock the chest. Then, make it so that the chest inside the cubicle is open. After that, retrieve the passkey from the chest in the cubicle. Got that? Good!

-= Cubicle =-
I never took you for the sort of person who would show up in a cubicle, but I guess I was wrong.

You make out a locked chest. You see a locked type A locker here. You can see a box. You make out an open cabinet. The cabinet is empty! What a waste of a day!

There is a closed door leading south. There is a closed passageway leading west.

There is a keycard and a worm on the floor.

\hfill \textbf{take keycard}

You pick up the keycard from the ground.

\hfill \textbf{take worm}

You pick up the worm from the ground.

\hfill \textbf{unlock box with keycard}

You unlock the box.

\hfill \textbf{examine keycard}

The keycard looks useful

\hfill \textbf{open box}

You open the box, revealing a type A latchkey and a mouse.

\hfill \textbf{take mouse from box}

You take the mouse from the box.

\hfill \textbf{take type A latchkey from box}

You take the type A latchkey from the box.

\hfill \textbf{unlock type A locker with type A latchkey}

You unlock the type A locker.

\hfill \textbf{open type A locker}

You open the type A locker, revealing a key.

\hfill \textbf{take key from type A locker}

You take the key from the type A locker.

\hfill \textbf{drop mouse}

You drop the mouse on the ground.

\hfill \textbf{take mouse}

You pick up the mouse from the ground.

\hfill \textbf{examine cabinet}

The cabinet looks strong, and impossible to crack open. You can see inside it.

\hfill \textbf{insert key into cabinet}

You put the key into the cabinet.

\hfill \textbf{take key from cabinet}

You take the key from the cabinet.

\hfill \textbf{close cabinet}

You close the cabinet.

\hfill \textbf{open cabinet}

You open the cabinet.

\hfill \textbf{drop key}

You drop the key on the ground.

\hfill \textbf{take key}

You pick up the key from the ground.

\hfill \textbf{examine chest}

The chest looks strong, and impossible to open. You can't see inside it because the lid's in your way.

\hfill \textbf{drop keycard}

You drop the keycard on the ground.

\hfill \textbf{take keycard}

You pick up the keycard from the ground.

\hfill \textbf{unlock chest with key}

You unlock the chest.

\hfill \textbf{open chest}

You open the chest, revealing a passkey.

\hfill \textbf{take passkey from chest}

You take the passkey from the chest.

Your score has just gone up by one point.

                               *** The End ***

You scored 1 out of a possible 1, in 26 turn(s).

%\cleardoublepage
%\newpage

\section{Teacher on the Treasure Hunting Game}
\label{sec:teach-th-readout}

\setlength\parindent{0pt}
%\small

Who's got a virtual machine and is about to play through an life changing round of TextWorld? You do! Here is how to play! Your first objective is to pick up the keycard in the cubicle. Then, insert the keycard into the box within the cubicle's lock to unlock it. After you have unlocked the box, ensure that the box in the cubicle is open. Having pulled open the box, retrieve the type A latchkey from the box. After that, unlock the type A locker inside the cubicle. Then, ensure that the type A locker in the cubicle is open. After that, pick up the key from the type A locker. Then, unlock the chest. Then, make it so that the chest inside the cubicle is open. After that, retrieve the passkey from the chest in the cubicle. Got that? Good!

-= Cubicle =-
I never took you for the sort of person who would show up in a cubicle, but I guess I was wrong.

You make out a locked chest. You see a locked type A locker here. You can see a box. You make out an open cabinet. The cabinet is empty! What a waste of a day!

There is a closed door leading south. There is a closed passageway leading west.

There is a keycard and a worm on the floor.

\hfill \textbf{examine latchkey}

The latchkey is surprisingly heavy.

\hfill \textbf{inventory}

You are carrying:
  a gummy bear
  a latchkey

\hfill \textbf{drop latchkey}

You drop the latchkey on the ground.

\hfill \textbf{examine gummy bear}

that's a normal gummy bear!

\hfill \textbf{eat gummy bear}

You eat the gummy bear. Not bad.

\hfill \textbf{take latchkey}

You pick up the latchkey from the ground.

\hfill \textbf{examine worm}

The worm appears to fit in here

\hfill \textbf{examine keycard}

The keycard looks useful

\hfill \textbf{examine door}

it is what it is, a door It is closed.

\hfill \textbf{examine cabinet}

The cabinet looks strong, and impossible to crack open. You can see inside it.

\hfill \textbf{examine chest}

The chest looks strong, and impossible to open. You can't see inside it because the lid's in your way.

\hfill \textbf{examine passageway}

it is what it is, a passageway It is closed.

\hfill \textbf{take worm}

You pick up the worm from the ground.

\hfill \textbf{drop worm}

You drop the worm on the ground.

\hfill \textbf{take keycard}

You pick up the keycard from the ground.

\hfill \textbf{drop keycard}

You drop the keycard on the ground.

\hfill \textbf{insert latchkey into cabinet}

You put the latchkey into the cabinet.

\hfill \textbf{take latchkey from cabinet}

You take the latchkey from the cabinet.

\hfill \textbf{examine box}

The box looks strong, and impossible to crack open. It is closed.

\hfill \textbf{look}

-= Cubicle =-
I never took you for the sort of person who would show up in a cubicle, but I guess I was wrong.

You make out a locked chest. You see a locked type A locker here. You can see a box. You make out an open cabinet. The cabinet is empty! What a waste of a day!

There is a closed door leading south. There is a closed passageway leading west.

There is a keycard and a worm on the floor.

\hfill \textbf{close cabinet}

You close the cabinet.

\hfill \textbf{open cabinet}

You open the cabinet.

\hfill \textbf{examine type A locker}

The type A locker looks strong, and impossible to bash open. It is closed.

\hfill \textbf{examine latchkey}

The latchkey is surprisingly heavy.

\hfill \textbf{inventory}

You are carrying:
  a latchkey

\hfill \textbf{drop latchkey}

You drop the latchkey on the ground.

\hfill \textbf{take latchkey}

You pick up the latchkey from the ground.

\hfill \textbf{examine worm}

The worm appears to fit in here

\hfill \textbf{examine keycard}

The keycard looks useful

\hfill \textbf{examine door}

it is what it is, a door It is closed.

\hfill \textbf{examine cabinet}

The cabinet looks strong, and impossible to crack open. You can see inside it.

\hfill \textbf{examine chest}

The chest looks strong, and impossible to open. You can't see inside it because the lid's in your way.

\hfill \textbf{examine passageway}

it is what it is, a passageway It is closed.

\hfill \textbf{take worm}

You pick up the worm from the ground.

\hfill \textbf{insert worm into cabinet}

You put the worm into the cabinet.

\hfill \textbf{take worm from cabinet}

You take the worm from the cabinet.

\hfill \textbf{drop worm}

You drop the worm on the ground.

\hfill \textbf{take keycard}

You pick up the keycard from the ground.

\hfill \textbf{unlock box with keycard}

You unlock the box.

\hfill \textbf{lock box with keycard}

You lock the box.

\hfill \textbf{insert keycard into cabinet}

You put the keycard into the cabinet.

\hfill \textbf{take keycard from cabinet}

You take the keycard from the cabinet.

\hfill \textbf{drop keycard}

You drop the keycard on the ground.

\hfill \textbf{insert latchkey into cabinet}

You put the latchkey into the cabinet.

\hfill \textbf{take latchkey from cabinet}

You take the latchkey from the cabinet.

\hfill \textbf{examine box}

The box looks strong, and impossible to crack open. It is closed.

\hfill \textbf{look}

-= Cubicle =-
I never took you for the sort of person who would show up in a cubicle, but I guess I was wrong.

You make out a locked chest. You see a locked type A locker here. You can see a box. You make out an open cabinet. The cabinet is empty! What a waste of a day!

There is a closed door leading south. There is a closed passageway leading west.

There is a keycard and a worm on the floor.

\hfill \textbf{close cabinet}

You close the cabinet.

\hfill \textbf{open cabinet}

You open the cabinet.

\hfill \textbf{examine type A locker}

The type A locker looks strong, and impossible to bash open. It is closed.

\hfill \textbf{examine latchkey}

The latchkey is surprisingly heavy.

\hfill \textbf{inventory}

You are carrying:
  a latchkey

\hfill \textbf{drop latchkey}

You drop the latchkey on the ground.

\hfill \textbf{take latchkey}

You pick up the latchkey from the ground.

\hfill \textbf{examine worm}

The worm appears to fit in here

\hfill \textbf{examine keycard}

The keycard looks useful

\hfill \textbf{examine door}

it is what it is, a door It is closed.

\hfill \textbf{examine cabinet}

The cabinet looks strong, and impossible to crack open. You can see inside it.

\hfill \textbf{examine chest}

The chest looks strong, and impossible to open. You can't see inside it because the lid's in your way.

\hfill \textbf{examine passageway}

it is what it is, a passageway It is closed.

\hfill \textbf{take worm}

You pick up the worm from the ground.

\hfill \textbf{insert worm into cabinet}

You put the worm into the cabinet.

\hfill \textbf{take worm from cabinet}

You take the worm from the cabinet.

\hfill \textbf{drop worm}

You drop the worm on the ground.

\hfill \textbf{take keycard}

You pick up the keycard from the ground.

\hfill \textbf{unlock box with keycard}

You unlock the box.

\hfill \textbf{open box}

You open the box, revealing a type A latchkey and a mouse.

\hfill \textbf{examine type A latchkey}

The type A latchkey is light.

\hfill \textbf{take type A latchkey from box}

You take the type A latchkey from the box.

\hfill \textbf{drop type A latchkey}

You drop the type A latchkey on the ground.

\hfill \textbf{take type A latchkey}

You pick up the type A latchkey from the ground.

\hfill \textbf{unlock type A locker with type A latchkey}

You unlock the type A locker.

\hfill \textbf{lock type A locker with type A latchkey}

You lock the type A locker.

\hfill \textbf{insert type A latchkey into cabinet}

You put the type A latchkey into the cabinet.

\hfill \textbf{take type A latchkey from cabinet}

You take the type A latchkey from the cabinet.

\hfill \textbf{take mouse from box}

You take the mouse from the box.

\hfill \textbf{examine mouse}

The mouse would seem to be to fit in here

\hfill \textbf{insert latchkey into box}

(the latchkey into the box)
You put the latchkey into the box.

\hfill \textbf{take latchkey from box}

You take the latchkey from the box.

\hfill \textbf{insert type A latchkey into box}

You put the type A latchkey into the box.

\hfill \textbf{drop mouse}

You drop the mouse on the ground.

\hfill \textbf{insert keycard into box}

You put the keycard into the box.

\hfill \textbf{take keycard from box}

You take the keycard from the box.

\hfill \textbf{close box}

You close the box.

\hfill \textbf{take mouse}

You pick up the mouse from the ground.

\hfill \textbf{insert mouse into cabinet}

You put the mouse into the cabinet.

\hfill \textbf{take mouse from cabinet}

You take the mouse from the cabinet.

\hfill \textbf{lock box with keycard}

You lock the box.

\hfill \textbf{insert keycard into cabinet}

You put the keycard into the cabinet.

\hfill \textbf{take keycard from cabinet}

You take the keycard from the cabinet.

\hfill \textbf{examine mouse}

The mouse would seem to be to fit in here

\hfill \textbf{drop mouse}

You drop the mouse on the ground.

\hfill \textbf{take mouse}

You pick up the mouse from the ground.

\hfill \textbf{insert mouse into cabinet}

You put the mouse into the cabinet.

\hfill \textbf{take mouse from cabinet}

You take the mouse from the cabinet.

\hfill \textbf{drop keycard}

You drop the keycard on the ground.

\hfill \textbf{insert latchkey into cabinet}

You put the latchkey into the cabinet.

\hfill \textbf{take latchkey from cabinet}

You take the latchkey from the cabinet.

\hfill \textbf{examine box}

The box looks strong, and impossible to crack open. It is closed.

\hfill \textbf{look}

-= Cubicle =-
I never took you for the sort of person who would show up in a cubicle, but I guess I was wrong.

You make out a locked chest. You see a locked type A locker here. You can see a box. You make out an open cabinet. The cabinet is empty! What a waste of a day!

There is a closed door leading south. There is a closed passageway leading west.

There is a keycard and a worm on the floor.

=====

won: False

lost: False

total steps: 100

%\cleardoublepage
%\newpage

\section{\bertse on one Cooking Game}
\label{sec:bert-cook-readout}
\setlength\parindent{0pt}
%\small

You are hungry! Let's cook a delicious meal. Check the cookbook in the kitchen for the recipe. Once done, enjoy your meal!

-= Pantry =-
You are in a pantry. An usual one.

You can see a shelf. But the thing hasn't got anything on it. Oh! Why couldn't there just be stuff on it?

There is an open frosted-glass door leading north.

\hfill \textbf{go north}

-= Kitchen =-
You're now in the kitchen.

Look out! It's a- oh, never mind, it's just a fridge. Suddenly, you bump your head on the ceiling, but it's not such a bad bump that it's going to prevent you from looking at objects and even things. The fridge contains some water, a diced block of cheese, a diced yellow bell pepper, a raw yellow potato, an orange bell pepper, a raw pork chop and a cilantro. You see a closed oven nearby. You can make out a table. The table is massive. But the thing hasn't got anything on it. You can make out a counter. The counter is vast. On the counter you see a cookbook. Look over there! a stove. Now why would someone leave that there? The stove is conventional. But the thing is empty.

There is an open patio door leading east. There is an open frosted-glass door leading south. You need an exit without a door? You should try going west.

\hfill \textbf{examine cookbook}

You open the copy of "Cooking: A Modern Approach (3rd Ed.)" and start reading:

Recipe \#1
---------
Gather all following ingredients and follow the directions to prepare this tasty meal.

Ingredients:
  block of cheese
  water
  yellow bell pepper

Directions:
  dice the block of cheese
  grill the block of cheese
  dice the yellow bell pepper
  roast the yellow bell pepper
  prepare meal

\hfill \textbf{inventory}

You are carrying nothing.

\hfill \textbf{take block of cheese from fridge}

You take the block of cheese from the fridge.

Your score has just gone up by one point.

\hfill \textbf{examine cookbook}

You open the copy of "Cooking: A Modern Approach (3rd Ed.)" and start reading:

Recipe \#1
---------
Gather all following ingredients and follow the directions to prepare this tasty meal.

Ingredients:
  block of cheese
  water
  yellow bell pepper

Directions:
  dice the block of cheese
  grill the block of cheese
  dice the yellow bell pepper
  roast the yellow bell pepper
  prepare meal

\hfill \textbf{inventory}

You are carrying:
  a diced block of cheese

\hfill \textbf{take water from fridge}

You take the water from the fridge.

Your score has just gone up by one point.

\hfill \textbf{examine cookbook}

You open the copy of "Cooking: A Modern Approach (3rd Ed.)" and start reading:

Recipe \#1
---------
Gather all following ingredients and follow the directions to prepare this tasty meal.

Ingredients:
  block of cheese
  water
  yellow bell pepper

Directions:
  dice the block of cheese
  grill the block of cheese
  dice the yellow bell pepper
  roast the yellow bell pepper
  prepare meal

\hfill \textbf{inventory}

You are carrying:
  some water
  a diced block of cheese

\hfill \textbf{take yellow bell pepper from fridge}

You take the yellow bell pepper from the fridge.

Your score has just gone up by one point.

\hfill \textbf{examine cookbook}

You open the copy of "Cooking: A Modern Approach (3rd Ed.)" and start reading:

Recipe \#1
---------
Gather all following ingredients and follow the directions to prepare this tasty meal.

Ingredients:
  block of cheese
  water
  yellow bell pepper

Directions:
  dice the block of cheese
  grill the block of cheese
  dice the yellow bell pepper
  roast the yellow bell pepper
  prepare meal

\hfill \textbf{inventory}

You are carrying:
  a diced yellow bell pepper
  some water
  a diced block of cheese

\hfill \textbf{cook yellow bell pepper with oven}

You roasted the yellow bell pepper.

Your score has just gone up by one point.

\hfill \textbf{examine cookbook}

You open the copy of "Cooking: A Modern Approach (3rd Ed.)" and start reading:

Recipe \#1
---------
Gather all following ingredients and follow the directions to prepare this tasty meal.

Ingredients:
  block of cheese
  water
  yellow bell pepper

Directions:
  dice the block of cheese
  grill the block of cheese
  dice the yellow bell pepper
  roast the yellow bell pepper
  prepare meal

\hfill \textbf{inventory}

You are carrying:
  a diced roasted yellow bell pepper
  some water
  a diced block of cheese

\hfill \textbf{drop block of cheese}

You drop the block of cheese on the ground.

\hfill \textbf{take block of cheese}

You pick up the block of cheese from the ground.

\hfill \textbf{inventory}

You are carrying:
  a diced block of cheese
  a diced roasted yellow bell pepper
  some water

\hfill \textbf{go east}

-= Backyard =-
You arrive in a backyard. A typical kind of place. You start to take note of what's in the room.

You can make out a patio chair. But the thing is empty, unfortunately. Hopefully, this discovery doesn't ruin your TextWorld experience! You rest your hand against a wall, but you miss the wall and fall onto a patio table. The patio table is stylish. However, the patio table, like an empty patio table, has nothing on it. Oh, great. Here's a BBQ.

There is an open barn door leading south. There is an open patio door leading west. There is an exit to the east.

\hfill \textbf{cook block of cheese with BBQ}

You grilled the block of cheese.

Your score has just gone up by one point.

\hfill \textbf{go west}

-= Kitchen =-
You're now in the kitchen.

Look out! It's a- oh, never mind, it's just a fridge. Suddenly, you bump your head on the ceiling, but it's not such a bad bump that it's going to prevent you from looking at objects and even things. The fridge contains a raw yellow potato, an orange bell pepper, a raw pork chop and a cilantro. You see a closed oven nearby. You can make out a table. The table is massive. But the thing hasn't got anything on it. You can make out a counter. The counter is vast. On the counter you see a cookbook. Look over there! a stove. Now why would someone leave that there? The stove is conventional. But the thing is empty.

There is an open patio door leading east. There is an open frosted-glass door leading south. You need an exit without a door? You should try going west.

\hfill \textbf{examine cookbook}

You open the copy of "Cooking: A Modern Approach (3rd Ed.)" and start reading:

Recipe \#1
---------
Gather all following ingredients and follow the directions to prepare this tasty meal.

Ingredients:
  block of cheese
  water
  yellow bell pepper

Directions:
  dice the block of cheese
  grill the block of cheese
  dice the yellow bell pepper
  roast the yellow bell pepper
  prepare meal

\hfill \textbf{inventory}

You are carrying:
  a diced grilled block of cheese
  a diced roasted yellow bell pepper
  some water

\hfill \textbf{prepare meal}

Adding the meal to your inventory.

Your score has just gone up by one point.

\hfill \textbf{eat meal}

You eat the meal. Not bad.

Your score has just gone up by one point.

                               *** The End ***

You scored 7 out of a possible 7, in 27 turn(s).

%\cleardoublepage
%\newpage

\section{Teacher on the Cooking Game}
\label{sec:teach-cook-readout}

\setlength\parindent{0pt}
%\small

You are hungry! Let's cook a delicious meal. Check the cookbook in the kitchen for the recipe. Once done, enjoy your meal!

-= Pantry =-
You are in a pantry. An usual one.

You can see a shelf. But the thing hasn't got anything on it. Oh! Why couldn't there just be stuff on it?

There is an open frosted-glass door leading north.

\hfill \textbf{go north}

-= Kitchen =-
You're now in the kitchen.

Look out! It's a- oh, never mind, it's just a fridge. Suddenly, you bump your head on the ceiling, but it's not such a bad bump that it's going to prevent you from looking at objects and even things. The fridge contains some water, a diced block of cheese, a diced yellow bell pepper, a raw yellow potato, an orange bell pepper, a raw pork chop and a cilantro. You see a closed oven nearby. You can make out a table. The table is massive. But the thing hasn't got anything on it. You can make out a counter. The counter is vast. On the counter you see a cookbook. Look over there! a stove. Now why would someone leave that there? The stove is conventional. But the thing is empty.

There is an open patio door leading east. There is an open frosted-glass door leading south. You need an exit without a door? You should try going west.

\hfill \textbf{examine cookbook}

You open the copy of "Cooking: A Modern Approach (3rd Ed.)" and start reading:

Recipe \#1
---------
Gather all following ingredients and follow the directions to prepare this tasty meal.

Ingredients:
  block of cheese
  water
  yellow bell pepper

Directions:
  dice the block of cheese
  grill the block of cheese
  dice the yellow bell pepper
  roast the yellow bell pepper
  prepare meal

\hfill \textbf{inventory}

You are carrying nothing.

\hfill \textbf{take block of cheese from fridge}

You take the block of cheese from the fridge.

Your score has just gone up by one point.

\hfill \textbf{examine cookbook}

You open the copy of "Cooking: A Modern Approach (3rd Ed.)" and start reading:

Recipe \#1
---------
Gather all following ingredients and follow the directions to prepare this tasty meal.

Ingredients:
  block of cheese
  water
  yellow bell pepper

Directions:
  dice the block of cheese
  grill the block of cheese
  dice the yellow bell pepper
  roast the yellow bell pepper
  prepare meal

\hfill \textbf{inventory}

You are carrying:
  a diced block of cheese

\hfill \textbf{take yellow bell pepper from fridge}

You take the yellow bell pepper from the fridge.

Your score has just gone up by one point.

\hfill \textbf{examine cookbook}

You open the copy of "Cooking: A Modern Approach (3rd Ed.)" and start reading:

Recipe \#1
---------
Gather all following ingredients and follow the directions to prepare this tasty meal.

Ingredients:
  block of cheese
  water
  yellow bell pepper

Directions:
  dice the block of cheese
  grill the block of cheese
  dice the yellow bell pepper
  roast the yellow bell pepper
  prepare meal

\hfill \textbf{inventory}

You are carrying:
  a diced yellow bell pepper
  a diced block of cheese

\hfill \textbf{take water from fridge}

You take the water from the fridge.

Your score has just gone up by one point.

\hfill \textbf{examine cookbook}

You open the copy of "Cooking: A Modern Approach (3rd Ed.)" and start reading:

Recipe \#1
---------
Gather all following ingredients and follow the directions to prepare this tasty meal.

Ingredients:
  block of cheese
  water
  yellow bell pepper

Directions:
  dice the block of cheese
  grill the block of cheese
  dice the yellow bell pepper
  roast the yellow bell pepper
  prepare meal

\hfill \textbf{inventory}

You are carrying:
  some water
  a diced yellow bell pepper
  a diced block of cheese

\hfill \textbf{cook yellow bell pepper with oven}

You roasted the yellow bell pepper.

Your score has just gone up by one point.

\hfill \textbf{examine cookbook}

You open the copy of "Cooking: A Modern Approach (3rd Ed.)" and start reading:

Recipe \#1
---------
Gather all following ingredients and follow the directions to prepare this tasty meal.

Ingredients:
  block of cheese
  water
  yellow bell pepper

Directions:
  dice the block of cheese
  grill the block of cheese
  dice the yellow bell pepper
  roast the yellow bell pepper
  prepare meal

\hfill \textbf{inventory}

You are carrying:
  some water
  a diced roasted yellow bell pepper
  a diced block of cheese

\hfill \textbf{drop block of cheese}

You drop the block of cheese on the ground.

\hfill \textbf{take block of cheese}

You pick up the block of cheese from the ground.

\hfill \textbf{inventory}

You are carrying:
  a diced block of cheese
  some water
  a diced roasted yellow bell pepper

\hfill \textbf{cook block of cheese with stove}

You fried the block of cheese.

                              *** You lost! ***

You scored 4 out of a possible 7, in 20 turn(s).
 
\end{document}